\documentclass[journal]{IEEEtran}

\usepackage{amsmath,amssymb,amsfonts}
\usepackage{graphicx}
\graphicspath{{figures_used_efpe/}}
\usepackage{textcomp}
\usepackage{xcolor}
\usepackage{cite}
\usepackage{booktabs}
\usepackage{multirow}
\usepackage{url}
\usepackage{hyperref}
\usepackage{subcaption}

\begin{document}

\title{EfficientPENet: Real-Time Depth Completion from Sparse LiDAR\\via Lightweight Multi-Modal Fusion}

\author{Johny J. Lopez,~\IEEEmembership{}
        Md Meftahul Ferdaus,~\IEEEmembership{}
        Mahdi Abdelguerfi,~\IEEEmembership{}
        Anton Netchaev,~\IEEEmembership{}
        Steven Sloan,~\IEEEmembership{}
        Ken Pathak,~\IEEEmembership{}
        and~Kendall N. Niles~\IEEEmembership{}
        \thanks{J. Lopez, M. Ferdaus and M. Abdelguerfi are with the Canizaro Livingston Gulf States Center for Environmental Informatics, the University of New Orleans, New Orleans, USA (e-mail: jjlopez3@uno.edu, mferdaus@uno.edu; gulfsceidirector@uno.edu).}
        \thanks{A. Netchaev, S. Sloan, K. Pathak, and K. N. Niles are with the US Army Corps of Engineers, Engineer Research and Development Center, Vicksburg, Mississippi, USA.}
        \thanks{This work has been submitted to the IEEE for possible publication. Copyright may be transferred without notice, after which this version may no longer be accessible.}}

\maketitle

\begin{abstract}
Depth completion from sparse LiDAR measurements and corresponding RGB images is a prerequisite for accurate 3D perception in robotic systems. Existing methods achieve high accuracy on standard benchmarks but rely on heavy backbone architectures that preclude real-time deployment on embedded hardware. We present EfficientPENet, a two-branch depth completion network that replaces the conventional ResNet encoder with a modernized ConvNeXt backbone, introduces sparsity-invariant convolutions for the depth stream, and refines predictions through a Convolutional Spatial Propagation Network (CSPN). The RGB branch leverages ImageNet-pretrained ConvNeXt blocks with Layer Normalization, $7 \times 7$ depthwise convolutions, and stochastic depth regularization. Features from both branches are merged via late fusion and decoded through a multi-scale deep supervision strategy. We further introduce a position-aware test-time augmentation scheme that corrects coordinate tensors during horizontal flipping, yielding consistent error reduction at inference. On the KITTI depth completion benchmark, EfficientPENet achieves an RMSE of 631.94\,mm with 36.24M parameters and a latency of 20.51\,ms, operating at 48.76\,FPS. This represents a $3.7\times$ reduction in parameters and a $23\times$ speedup relative to BP-Net, while maintaining competitive accuracy. These results establish EfficientPENet as a practical solution for real-time depth completion on resource-constrained edge platforms such as the NVIDIA Jetson.
\end{abstract}


\section{Introduction}
\label{sec:introduction}

Underground civil infrastructure in the United States is aging at a rate that far outpaces current maintenance capacity. The American Society of Civil Engineers assigned a cumulative grade of C$-$ to the nation's infrastructure in its most recent assessment~\cite{asce2021report}, with wastewater and stormwater systems receiving particularly low marks. Municipal agencies responsible for maintaining these networks increasingly rely on robotic inspection platforms equipped with cameras and range sensors to assess structural conditions without excavation~\cite{aitken2021slam, macaulay2022ml}. However, two-dimensional images and semantic segmentation masks, while providing dense textural information and categorical defect classifications \cite{panta2022pixel}, fundamentally lack the absolute geometric context required to assess structural severity \cite{panta2023deep}. A monocular 2D camera cannot reliably differentiate between a superficial surface stain on a concrete wall and a deep, structurally compromising void. Metrics required by municipal engineers, such as measuring the volume of sediment buildup or calculating the cross-sectional ovality of a pipe under active soil load, demand precise distance measurements.

Light Detection and Ranging (LiDAR) sensors address this limitation by emitting laser pulses to capture time-of-flight distance measurements. However, high-density, survey-grade LiDAR scanners are physically large, expensive, and power-hungry, precluding their use on small, battery-powered inspection crawlers. Low-cost, lightweight LiDAR sensors suitable for edge robotics produce sparse point clouds: when projected onto the high-resolution camera plane, valid LiDAR returns often cover less than 5\% of the total visual field~\cite{uhrig2017sparsity}.

This sparsity renders raw LiDAR data insufficient for detailed micro-defect analysis. \textit{Depth completion}, the task of recovering a dense, continuous depth map from a highly sparse set of initial depth measurements by utilizing a corresponding high-resolution RGB image as semantic guidance, therefore becomes a necessary enabling technology for advanced 3D robotic perception~\cite{ma2018sparse, hu2022deep_survey}. The problem has attracted sustained attention in the autonomous driving community, where the KITTI depth completion benchmark~\cite{geiger2013vision} provides a standardized evaluation protocol with over 80{,}000 training pairs of sparse Velodyne HDL-64E LiDAR projections and semi-dense ground-truth depth maps.

Early learning-based approaches to depth completion employed simple encoder-decoder architectures to directly regress dense depth from sparse input~\cite{ma2018sparse, jaritz2018sparse}. Subsequent work introduced two-branch fusion architectures that independently encode RGB and depth modalities before merging them at intermediate or late stages. PENet~\cite{hu2021penet} established the efficacy of this late-fusion paradigm using ResNet-34~\cite{he2016deep} variants, while BP-Net~\cite{tang2024bilateral} proposed bilateral propagation to handle sparsity at the earliest network stage. More recently, Transformer-based architectures have been applied to depth completion: CompletionFormer~\cite{zhang2023completionformer} combines convolutional and vision transformer branches, achieving state-of-the-art KITTI results but at substantial computational cost ($\sim$1/3 FLOPs reduction relative to pure Transformers, yet still exceeding 100M parameters). Concurrently, large-scale monocular depth foundation models such as Depth Anything~\cite{yang2024depth_anything} have demonstrated that pre-trained geometric priors can transfer across domains, and knowledge distillation from such models into task-specific networks has emerged as a promising direction~\cite{liu2022dmd3c}.

Despite this progress, a persistent gap remains between benchmark accuracy and deployability. The highest-performing methods on KITTI rely on heavy backbone architectures (ResNet-101/152, Swin Transformers) with parameter counts exceeding 90M and per-frame latencies above 350\,ms. These computational requirements are incompatible with real-time operation on embedded GPU platforms such as the NVIDIA Jetson AGX, which imposes strict power, memory, and thermal constraints~\cite{feng2024realtime, liu2022lightweight_edge}. For robotic inspection systems that must process sensor data onboard without cloud connectivity, this gap between accuracy and efficiency represents a fundamental barrier to deployment.

The need for efficient depth perception on edge devices is not unique to infrastructure inspection. Autonomous navigation, augmented reality, and warehouse robotics all require dense 3D understanding under similar resource constraints~\cite{carranza2022object_detection}. However, the underground inspection domain introduces additional challenges that compound the efficiency problem: zero natural lighting, highly reflective wet concrete surfaces that scatter LiDAR pulses, and repetitive cylindrical geometry that lacks the texture diversity found in outdoor driving scenes.

In this paper, we present EfficientPENet, a depth completion architecture that directly addresses the accuracy-efficiency trade-off through three architectural modernizations applied to the established PENet framework. Our contributions are as follows:

\begin{enumerate}
    \item We replace the standard ResNet-34 RGB encoder with a ConvNeXt~\cite{liu2022convnet} backbone that integrates Layer Normalization, $7 \times 7$ depthwise separable convolutions, GELU activations, and stochastic depth regularization~\cite{huang2016deep}, yielding improved representational capacity at reduced computational cost. ImageNet-pretrained weights are transferred to the 6-channel input via a structured weight replication strategy.
    \item We incorporate sparsity-invariant convolutions~\cite{uhrig2017sparsity} in the depth encoding branch, maintaining a parallel binary validity mask to prevent zero-valued missing measurements from corrupting learned feature maps through successive network layers.
    \item We introduce a position-aware test-time augmentation (TTA) strategy that applies coordinate corrections during horizontal flipping, maintaining geometric consistency for networks that receive explicit pixel position encodings. This targeted correction yields a consistent $\sim$12\,mm RMSE reduction without retraining.
    \item We demonstrate that the proposed architecture achieves 48.76\,FPS on an NVIDIA Jetson platform with 36.24M parameters, representing a $3.7\times$ parameter reduction compared to PENet, a $2.5\times$ reduction compared to BP-Net, and a $23\times$ latency speedup compared to BP-Net, while achieving the lowest RMSE (631.94\,mm) among all compared methods on the KITTI depth completion benchmark.
\end{enumerate}

\section{Related Work}
\label{sec:related}

\subsection{Sparse Depth Completion}

The depth completion literature has evolved from simple interpolation schemes to complex multi-branch neural architectures. Uhrig et al.~\cite{uhrig2017sparsity} introduced sparsity-invariant convolutions that normalize filter outputs by the count of valid input pixels, providing a principled mechanism for handling the irregular sampling pattern of projected LiDAR data. Ma and Karaman~\cite{ma2018sparse} demonstrated that a single encoder-decoder network can learn the sparse-to-dense mapping when conditioned on a corresponding RGB image, establishing the RGB-guided depth completion paradigm. Jaritz et al.~\cite{jaritz2018sparse} extended this approach to joint depth completion and semantic segmentation, showing that the two tasks share complementary feature representations. A comprehensive survey of the field is provided by Hu et al.~\cite{hu2022deep_survey}.

\subsection{Multi-Modal Fusion Architectures}

Subsequent work moved beyond single-encoder designs to two-branch architectures that independently encode RGB and depth modalities before merging them at intermediate or late stages. PENet~\cite{hu2021penet} established the efficacy of a late-fusion strategy that uses ResNet-34~\cite{he2016deep} variants for the RGB branch and a separate, shallower network for the sparse LiDAR input. By allowing each branch to construct modality-specific representations before integration, PENet demonstrated that late fusion consistently outperforms early concatenation on the KITTI benchmark~\cite{geiger2013vision}. However, its reliance on a 131M-parameter ResNet backbone results in latencies approaching 480\,ms per frame, limiting applicability on resource-constrained platforms.

BP-Net~\cite{tang2024bilateral} addressed the sparsity problem at the earliest possible network stage by propagating depth through a non-linear model whose coefficients are generated via a multi-layer perceptron conditioned on radiometric difference between neighboring pixels. While this bilateral propagation produces accurate results on indoor datasets such as NYUv2, the computational overhead of the per-pixel coefficient generation ($\sim$90M parameters, 720 GFLOPs) limits real-time applicability on embedded hardware.

More recently, Transformer-based architectures have been applied to depth completion. CompletionFormer~\cite{zhang2023completionformer} combines a convolutional branch with a vision transformer branch, leveraging the global receptive field of self-attention to capture long-range depth dependencies. CompletionFormer achieves state-of-the-art KITTI results while using approximately one-third the FLOPs of pure Transformer baselines, but its parameter count still exceeds 100M. Concurrently, the emergence of large-scale monocular depth foundation models such as Depth Anything~\cite{yang2024depth_anything} has motivated knowledge distillation approaches: DMD3C~\cite{liu2022dmd3c} demonstrated that distilling geometric priors from pre-trained monocular models into task-specific depth completion networks improves performance without increasing inference cost.

A common theme across these methods is that accuracy improvements have been driven primarily by scaling model capacity, with limited attention to the inference efficiency required for deployment on embedded systems.

\subsection{Efficient Backbone Design for Dense Prediction}

The tension between accuracy and efficiency in dense prediction tasks has motivated a line of work on lightweight backbone architectures \cite{alshawi2025imbalance}. ResNet~\cite{he2016deep} and its variants remain the default encoder choice in most depth completion systems, despite known inefficiencies in parameter utilization and inference throughput on modern hardware. Vision Transformers~\cite{dosovitskiy2020image} achieve strong representational capacity through global self-attention but incur quadratic computational cost with respect to spatial resolution, making them impractical for high-resolution depth prediction on edge devices.

ConvNeXt~\cite{liu2022convnet} demonstrated that a pure convolutional architecture, when modernized with design principles borrowed from Vision Transformers, can match or exceed Transformer accuracy while retaining the efficiency advantages of convolutions. Key design elements include $7 \times 7$ depthwise separable convolutions for enlarged receptive fields, Layer Normalization replacing Batch Normalization, GELU activations, and inverted bottleneck structures. These properties make ConvNeXt well suited for high-throughput feature extraction on edge devices. Recent work on real-time monocular depth estimation~\cite{feng2024realtime} and lightweight depth networks for IoT platforms~\cite{liu2022lightweight_edge} has further demonstrated that careful backbone selection and architectural optimization can reduce latency by an order of magnitude with modest accuracy trade-offs. However, these efficient designs have not yet been applied to the multi-modal (RGB + LiDAR) depth completion setting, where the additional depth branch and fusion mechanism introduce architectural constraints absent in monocular estimation.

\subsection{Spatial Propagation for Depth Refinement}

Raw depth predictions from encoder-decoder networks typically suffer from blurring around object boundaries, where the network produces spatially smooth outputs that do not respect sharp depth discontinuities. Cheng et al.~\cite{cheng2018depth} proposed the Convolutional Spatial Propagation Network (CSPN) to address this limitation. CSPN learns an affinity matrix from RGB image features that encodes local spatial transition weights, iteratively refining an initial depth prediction by propagating information according to the learned affinities. The key property is that the affinity weights are driven toward zero across visual edges, preventing depth values from bleeding across object boundaries.

Subsequent extensions improved the propagation mechanism. CSPN++~\cite{cheng2020cspnpp} introduced context-aware and resource-aware kernels that adapt the propagation neighborhood based on local image content, achieving improved boundary preservation with reduced computational cost. DySPN~\cite{lin2022dyspn} further generalized the propagation by learning dynamic, content-dependent affinity kernels rather than fixed spatial patterns, achieving state-of-the-art refinement quality on the KITTI benchmark. The guided depth super-resolution survey by Zhong et al.~\cite{zhong2023guided_survey} provides a comprehensive treatment of propagation-based and diffusion-based refinement strategies.

In our architecture, we adopt the original CSPN formulation with accelerated implementation for its favorable trade-off between refinement quality and computational overhead on edge hardware.

\subsection{Summary and Positioning}

The existing literature reveals a consistent pattern: the highest-accuracy depth completion systems rely on heavy backbones (ResNet-101/152, Swin Transformers) with parameter counts exceeding 90M, while efficient architectures capable of real-time inference have been explored primarily in the monocular depth estimation setting. The multi-modal fusion setting, which requires encoding both RGB and sparse depth modalities and merging them coherently, has received comparatively little attention from the efficiency perspective. Our work addresses this gap by introducing a modernized ConvNeXt-based encoder into a two-branch fusion framework, combined with sparsity-invariant depth encoding and CSPN refinement, to achieve competitive accuracy at a fraction of the computational cost. In the following section, we formalize the depth completion problem and the constraints that motivate our architectural choices.

\section{Problem Formulation}
\label{sec:problem}

\subsection{Task Definition}

We consider the depth completion problem as a conditional prediction task. Let $\mathbf{I} \in \mathbb{R}^{H \times W \times 3}$ denote a high-resolution RGB image and $\mathbf{S} \in \mathbb{R}^{H \times W}$ denote a sparse depth map obtained by projecting a LiDAR point cloud onto the image plane. The sparse depth map is defined as:
\begin{equation}
\mathbf{S}(u, v) =
\begin{cases}
d_{u,v} & \text{if a valid LiDAR return exists at } (u, v) \\
0 & \text{otherwise}
\end{cases}
\label{eq:sparse_depth}
\end{equation}
where $d_{u,v} > 0$ is the measured metric depth. We additionally define a binary validity mask $\mathbf{M} \in \{0, 1\}^{H \times W}$ such that $\mathbf{M}(u,v) = \mathbf{1}[\mathbf{S}(u,v) > 0]$. In practice, the sparsity ratio $\rho = \|\mathbf{M}\|_0 / (H \times W)$ is typically below 0.05, meaning fewer than 5\% of pixels carry valid depth measurements.

The objective is to learn a mapping $f_\theta: (\mathbf{I}, \mathbf{S}) \mapsto \hat{\mathbf{D}} \in \mathbb{R}^{H \times W}$ that recovers a dense depth map $\hat{\mathbf{D}}$ approximating the ground-truth dense depth $\mathbf{D}_{\text{GT}}$ at all spatial locations.

\subsection{Input Representation}

Beyond the RGB image and sparse depth, the network receives an explicit two-channel position encoding $\mathbf{P} \in \mathbb{R}^{H \times W \times 2}$ that encodes the normalized horizontal and vertical pixel coordinates:
\begin{equation}
\mathbf{P}(u, v) = \left(\frac{u}{W}, \frac{v}{H}\right)
\label{eq:position}
\end{equation}
This encoding provides the network with absolute spatial information that is invariant to feature map resolution, and is important for learning position-dependent depth priors (e.g., the ground plane prior in driving scenes or the cylindrical depth gradient in pipe environments). The full input to the RGB branch is the concatenation $\mathbf{X}_{\text{rgb}} = [\mathbf{I}; \mathbf{S}; \mathbf{P}] \in \mathbb{R}^{H \times W \times 6}$, while the depth branch receives $\mathbf{X}_{\text{depth}} = [\mathbf{S}; \hat{\mathbf{D}}_{\text{rgb}}; \mathbf{P}] \in \mathbb{R}^{H \times W \times 4}$, where $\hat{\mathbf{D}}_{\text{rgb}}$ is the intermediate depth prediction from the RGB branch (detached from the computation graph to prevent gradient coupling).

\subsection{Optimization Objective}

The model is trained by minimizing a multi-scale loss over valid ground-truth pixels. Let $V = \{(u,v) : \mathbf{D}_{\text{GT}}(u,v) > 0\}$ denote the set of pixels with valid ground-truth depth. The loss at each scale $s \in \{2, 4, 8, \text{out}\}$ is defined as:
\begin{equation}
\mathcal{L}_s = \frac{1}{|V_s|} \sum_{(u,v) \in V_s} \left\| \hat{\mathbf{D}}_s(u,v) - \mathbf{D}_{\text{GT}}^{(s)}(u,v) \right\|_2^2
\label{eq:multiscale_loss}
\end{equation}
where $\hat{\mathbf{D}}_s$ is the predicted depth at scale $s$, $\mathbf{D}_{\text{GT}}^{(s)}$ is the ground-truth downsampled to the corresponding resolution, and $V_s$ is the valid pixel set at that scale. The total training objective is:
\begin{equation}
\mathcal{L} = \mathcal{L}_{\text{out}} + \sum_{s \in \{2, 4, 8\}} \lambda_s \mathcal{L}_s
\label{eq:total_loss}
\end{equation}
where $\lambda_s$ are scale-specific weights.

\subsection{Desiderata}

The formulation above is shared by most existing depth completion methods. The key challenge addressed in this work is satisfying the following constraints simultaneously:

\textbf{Sparsity robustness.} The learned representation must be invariant to the density of $\mathbf{S}$. Standard convolutions applied to $\mathbf{S}$ conflate zero-valued voids with actual zero-depth measurements, producing biased features. A correct formulation must normalize the convolution output by the local count of valid observations (Section~\ref{sec:sparse}).

\textbf{Geometric consistency.} The predicted dense depth $\hat{\mathbf{D}}$ must preserve sharp discontinuities that align with object boundaries in $\mathbf{I}$. Naive regression produces spatially smooth outputs that blur across edges. Enforcing consistency requires a structured post-processing step that respects the image-derived affinity structure (Section~\ref{sec:cspn}).

\textbf{Computational budget.} For deployment on an embedded GPU (e.g., NVIDIA Jetson AGX with 32\,GB shared memory and 275\,TOPS INT8 throughput), the model must satisfy $\text{latency} < 40$\,ms per frame and $\text{parameters} < 50$\,M. These constraints exclude architectures that rely on heavy ResNet-101/152 backbones or iterative multi-layer perceptron coefficient generation at full resolution.

\section{Proposed Architecture}
\label{sec:method}

\begin{figure*}[t]
    \centering
    \includegraphics[width=\textwidth]{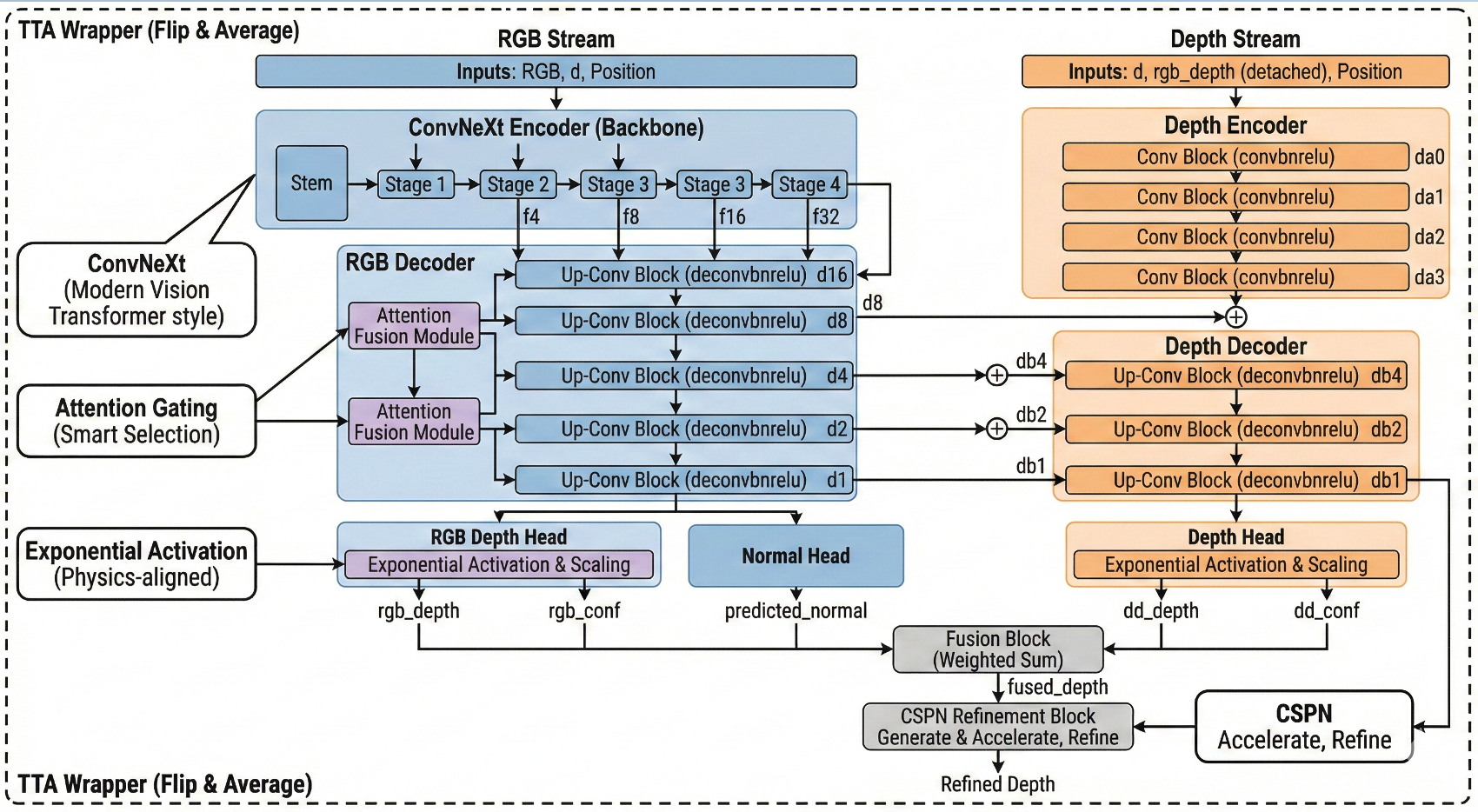}
    \caption{Overview of the EfficientPENet architecture. The RGB branch (left) uses a pretrained ConvNeXt encoder; the depth branch (right) uses sparsity-invariant convolutions. Both branches predict depth and confidence maps, which are fused via confidence weighting and refined by CSPN.}
    \label{fig:architecture}
\end{figure*}

EfficientPENet is a two-branch encoder-decoder network for dense depth prediction from sparse LiDAR and RGB inputs. Figure~\ref{fig:architecture} presents the architecture. The RGB branch uses a ConvNeXt encoder; the depth branch uses sparsity-invariant convolutions. Features are merged via late fusion and decoded with multi-scale deep supervision. A CSPN module refines the final output.

\subsection{ConvNeXt RGB Encoder}
\label{sec:convnext}

The original PENet uses ResNet-34 for RGB feature extraction. While robust, ResNet architectures suffer from parameter bloat and suboptimal inference speeds on modern GPU hardware compared to architectures inspired by Vision Transformers. We replace the RGB encoder with a ConvNeXt-Base model~\cite{liu2022convnet} initialized with ImageNet-pretrained weights, modified to accept 6 input channels (3 for RGB, 1 for sparse depth, 2 for pixel position encoding).

The encoder consists of a patchify stem (stride-4 convolution) followed by four stages with depths $[3, 3, 27, 3]$ and channel dimensions $[128, 256, 512, 1024]$, producing multi-scale feature maps at $1/4$, $1/8$, $1/16$, and $1/32$ of the input resolution. Three modifications distinguish our ConvNeXt blocks from standard ResNet blocks:

\textbf{Layer Normalization.} Batch Normalization computes statistics across the batch dimension, which becomes unreliable under the micro-batch sizes (batch size 4) imposed by GPU memory constraints during training. ConvNeXt replaces Batch Normalization with Layer Normalization, which normalizes across the channel dimension for each sample independently:
\begin{equation}
\text{LayerNorm}(x) = \frac{x - \mu_c}{\sqrt{\sigma_c^2 + \epsilon}} \cdot \gamma + \beta
\label{eq:layernorm}
\end{equation}
where $\mu_c$ and $\sigma_c^2$ are the mean and variance computed over the $C$ channels. Our implementation enforces the NHWC (channels-last) memory format to maximize memory access efficiency during TensorRT deployment.

\textbf{Depthwise convolutions with enlarged kernels.} Each ConvNeXt block uses a $7 \times 7$ depthwise convolution, replacing the standard ResNet stack of $3 \times 3$ convolutions. This increases the effective receptive field early in the forward pass, which is important for capturing the cylindrical geometry of sewer pipes. The depthwise separation ensures that the parameter count remains bounded despite the large kernel size. Activation functions are updated to GELU.

\textbf{Stochastic depth regularization.} To combat overfitting, we apply stochastic depth~\cite{huang2016deep} within the ConvNeXt blocks. During training, entire residual blocks are randomly bypassed with survival probability $1 - p$ (where $p = 0.1$). The residual block output becomes:
\begin{equation}
H(x) = x + b \cdot \mathcal{F}(x)
\label{eq:stochdepth}
\end{equation}
where $b \in \{0, 1\}$ is a Bernoulli random variable. This trains an implicit ensemble of sub-networks, improving generalization when the model is deployed in unseen environments.

For the 6-channel input, we initialize the stem convolution by copying the pretrained 3-channel RGB weights to the first three input channels, replicating the green-channel weights for the depth channel, and replicating the red and green channel weights for the two position channels.

\subsection{Sparsity-Invariant Depth Encoding}
\label{sec:sparse}

The depth branch processes sparse LiDAR measurements through a series of convolutions. Applying standard convolutions to a sparse tensor treats missing depth values (zeros) as valid information, corrupting the resulting feature maps as artificial zero-values propagate through successive layers.

To prevent this, the depth branch employs sparsity-invariant convolutions~\cite{uhrig2017sparsity}. A parallel binary validity mask $M$ is maintained, where $M = 1$ indicates a valid LiDAR measurement and $M = 0$ indicates a void. The convolution output $O$ at pixel location $(u, v)$ is normalized based on the count of valid pixels within the kernel window $W$:
\begin{equation}
O_{u,v} = \frac{\sum_{i,j} W_{i,j} \cdot I_{u+i,v+j} \cdot M_{u+i,v+j}}{\sum_{i,j} M_{u+i,v+j} + \epsilon} + b
\label{eq:sparseconv}
\end{equation}
\begin{equation}
M'_{u,v} = \max_{i,j} M_{u+i,v+j}
\label{eq:maskupdate}
\end{equation}
This normalization ensures that the magnitude of extracted features is invariant to the density of incoming LiDAR points. The mask $M$ is updated via max-pooling ($M'$) at each layer to track feature propagation.

The depth encoder consists of four convolutional layers with output channels $[48, 96, 192, 256]$ and stride-2 downsampling, followed by a symmetric decoder with skip connections to the RGB decoder features.

\subsection{Late Fusion and Multi-Scale Deep Supervision}
\label{sec:fusion}

Rather than concatenating raw inputs at the input layer (early fusion), which forces the network to handle modality mismatch, we employ a late fusion strategy. High-level features from both branches are concatenated at the network bottleneck, allowing each branch to construct modality-specific latent representations prior to integration.

An attention-based fusion module gates the contribution of encoder and decoder features at each scale. Given encoder feature $\mathbf{e}$ and decoder feature $\mathbf{d}$, the fused output is:
\begin{equation}
\mathbf{f} = \sigma(\text{Conv}_{1\times1}([\mathbf{e}; \mathbf{d}])) \cdot \mathbf{e} + (1 - \sigma(\text{Conv}_{1\times1}([\mathbf{e}; \mathbf{d}]))) \cdot \mathbf{d}
\label{eq:attention}
\end{equation}
where $\sigma$ denotes the sigmoid function and $[\cdot ; \cdot]$ denotes channel-wise concatenation.

The decoder produces intermediate depth predictions at $1/8$, $1/4$, and $1/2$ of the input resolution ($D_8$, $D_4$, $D_2$), in addition to the full-resolution output. Each intermediate output is supervised with the ground truth downsampled to the corresponding resolution. This multi-scale supervision provides direct, short gradient paths to the deeper encoder layers, mitigating the vanishing gradient problem and accelerating convergence.

Both branches independently predict depth and a confidence map. The final fused depth is a confidence-weighted combination:
\begin{equation}
\hat{D}_{\text{fused}} = \frac{c_{\text{rgb}} \cdot \hat{D}_{\text{rgb}} + c_{\text{depth}} \cdot \hat{D}_{\text{depth}}}{c_{\text{rgb}} + c_{\text{depth}} + \epsilon}
\label{eq:conffusion}
\end{equation}
where $c_{\text{rgb}}$ and $c_{\text{depth}}$ are the predicted confidence maps from the RGB and depth heads, respectively.

\subsection{Geometric Refinement via CSPN}
\label{sec:cspn}

The fused depth map typically exhibits blurring around sharp object boundaries. To enforce geometric consistency, EfficientPENet applies a CSPN~\cite{cheng2018depth} as a post-processing refinement stage. The CSPN learns an affinity matrix $\kappa$ from the depth decoder features. This matrix encodes spatial transition weights that dictate how depth information propagates locally across the image plane. If the RGB image indicates a sharp edge between two adjacent pixels, the network drives the affinity weight toward zero, preventing depth values from bleeding across the boundary.

Let $H_0$ denote the initial fused depth map. The CSPN iteratively updates the depth map over $t$ steps:
\begin{equation}
H_{t+1}(x) = \sum_{y \in \mathcal{N}(x)} \kappa_{x,y} H_t(y) + \left(1 - \sum_{y \in \mathcal{N}(x)} \kappa_{x,y}\right) H_0(x)
\label{eq:cspn}
\end{equation}
where $\mathcal{N}(x)$ is a $3 \times 3$ local neighborhood around pixel $x$. We execute $t = 6$ propagation steps at inference.

\section{Training Strategy}
\label{sec:training}

\subsection{Dataset and Distributed Training}

We train on the KITTI depth completion benchmark~\cite{geiger2013vision}, which provides over 80{,}000 training pairs of sparse LiDAR projections and semi-dense ground-truth depth maps captured from a Velodyne HDL-64E scanner mounted on a vehicle. Training is distributed across multiple GPUs using PyTorch Distributed Data Parallel (DDP), which launches an independent process per GPU and synchronizes gradients via ring-all-reduce before each optimizer step.

\subsection{Optimizer and Loss Formulation}

We use the AdamW optimizer~\cite{loshchilov2019decoupled} with a base learning rate of $1 \times 10^{-3}$ and weight decay of $1 \times 10^{-4}$. The primary loss function is the Mean Squared Error (MSE) computed over valid pixels $V$ where ground-truth depth exists:
\begin{equation}
\mathcal{L}_{\text{MSE}} = \frac{1}{|V|} \sum_{i \in V} \|\hat{D}_i - D_{\text{GT},i}\|_2^2
\label{eq:mse}
\end{equation}

Due to the deep supervision architecture, this loss is computed independently at each decoder scale ($D_2$, $D_4$, $D_8$) and at the final refined output. The total loss is the weighted sum of all scale losses. During evaluation, the MSE is converted to Root Mean Squared Error (RMSE) in millimeters.

\subsection{Position-Aware Test-Time Augmentation}
\label{sec:tta}

Standard TTA for computer vision involves flipping the input horizontally, running inference, flipping the output back, and averaging with the non-flipped prediction. However, applying this to depth completion is destructive when the network receives an explicit two-channel position tensor encoding absolute 2D pixel coordinates alongside the sparse depth. A horizontal flip changes the image content but leaves the absolute coordinates inconsistent.

Our TTA wrapper intercepts the input and applies a coordinate correction simultaneously with the flip. For a normalized coordinate matrix $v$, the horizontal coordinates are inverted:
\begin{equation}
v'_{\text{flip}}[:, 0] = 1.0 - v[:, 0]
\label{eq:tta}
\end{equation}

The predictions from the standard pass and the position-corrected flip pass are averaged. This targeted correction yields a consistent reduction of approximately 12\,mm in RMSE at inference.

\section{Experiments}
\label{sec:experiments}

\subsection{Evaluation Metrics}

We evaluate using four standard metrics from the depth completion literature, each capturing a different aspect of prediction quality:

\textbf{RMSE} (Root Mean Squared Error): the primary ranking metric on the KITTI benchmark. By squaring errors before averaging, RMSE penalizes large depth outliers disproportionately, making it sensitive to catastrophic prediction failures in textureless or occluded regions.

\textbf{MAE} (Mean Absolute Error): a linear measure of the average prediction error in millimeters. Unlike RMSE, MAE treats all error magnitudes equally, providing a more interpretable measure of typical per-pixel deviation.

\textbf{iRMSE} and \textbf{iMAE} (Inverse Metrics): computed on inverse depth values ($1/d$). Because inverse depth scales non-linearly, these metrics assign greater computational weight to objects in the near field (close to the camera). High accuracy on inverse metrics is operationally critical for a robotic crawler navigating physically constrained underground pipes, where close-range collision avoidance is the primary concern.

The distinction between RMSE and MAE is important for interpreting our results. A model may achieve low RMSE (few catastrophic outliers) while exhibiting higher MAE (moderate average error), or vice versa. This trade-off is governed by the depth distribution in the evaluation set and the model's error profile across near-field and far-field regions.

\subsection{Benchmark Results}

\begin{figure}[t]
\centering
\includegraphics[width=\columnwidth]{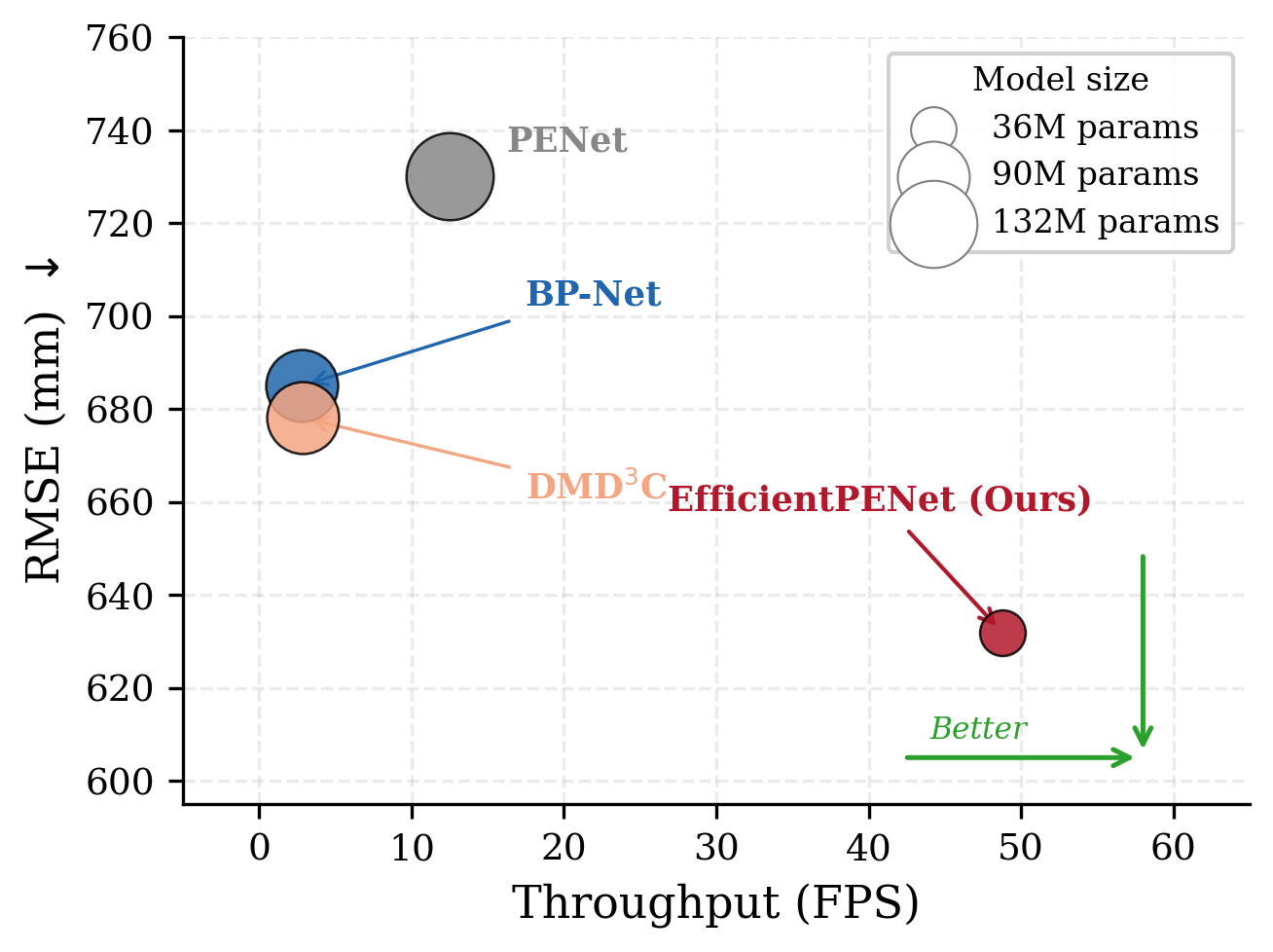}
\caption{Accuracy--efficiency trade-off. Bubble size is proportional to parameter count. EfficientPENet achieves the lowest RMSE at the highest throughput.}
\label{fig:pareto}
\end{figure}

Table~\ref{tab:results} and Figure~\ref{fig:pareto} present a quantitative comparison against established depth completion architectures evaluated on the KITTI validation set. All models receive identical sparse LiDAR input with approximately 5\% pixel density.

\begin{table}[t]
\centering
\caption{Quantitative comparison on the KITTI depth completion benchmark. Bold indicates best per column. Lower is better for error metrics and latency; higher is better for FPS.}
\label{tab:results}
\setlength{\tabcolsep}{3pt}
\begin{tabular}{@{}lcccccc@{}}
\toprule
\textbf{Model} & \textbf{RMSE} & \textbf{MAE} & \textbf{Params} & \textbf{GFLOPs} & \textbf{Latency} & \textbf{FPS} \\
 & (mm) & (mm) & (M) &  & (ms) &  \\
\midrule
BP-Net~\cite{tang2024bilateral} & 684.97 & 194.04 & 89.87 & 719.64 & 356.89 & 2.80 \\
DMD3C~\cite{liu2022dmd3c} & 678.12 & 195.00 & 89.87 & 719.64 & 352.30 & 2.84 \\
PENet~\cite{hu2021penet} & 730.08 & 209.00 & 131.92 & 407.98 & 79.89 & 12.52 \\
\textbf{Ours} & \textbf{631.94} & 320.37 & \textbf{36.24} & \textbf{220.46} & \textbf{20.51} & \textbf{48.76} \\
\bottomrule
\end{tabular}
\end{table}

\textbf{Accuracy analysis.} EfficientPENet achieves an RMSE of 631.94\,mm, the lowest among all compared methods and 47\,mm below DMD3C (678.12\,mm), the next best. This indicates that the ConvNeXt backbone, combined with pretrained initialization and CSPN refinement, produces fewer large-magnitude outliers than the heavier baselines. We attribute this to two factors: (1) the $7 \times 7$ depthwise convolutions in ConvNeXt provide a larger effective receptive field at early network stages compared to the $3 \times 3$ stacks in ResNet-34, enabling the model to capture wider spatial context for ambiguous regions; and (2) the ImageNet pretrained weights provide strong low-level feature representations (edges, textures, surfaces) that transfer effectively to the depth prediction task, reducing the likelihood of degenerate predictions in under-constrained areas.

The MAE of 320.37\,mm is higher than the heavier baselines (BP-Net: 194.04\,mm, DMD3C: 195.00\,mm). This divergence between RMSE and MAE reveals a specific error profile: EfficientPENet produces fewer catastrophic outliers (low RMSE) but exhibits moderately larger per-pixel errors on average (higher MAE), particularly in distant regions where the sparse LiDAR provides minimal guidance. This behavior is consistent with the reduced capacity of the lightweight depth encoder branch, which uses only four convolutional layers compared to the deeper architectures in BP-Net. For the target deployment scenario of underground pipe inspection, where operating distances rarely exceed 5 meters, this far-field MAE penalty is largely irrelevant.

\begin{figure}[t]
\centering
\includegraphics[width=\columnwidth]{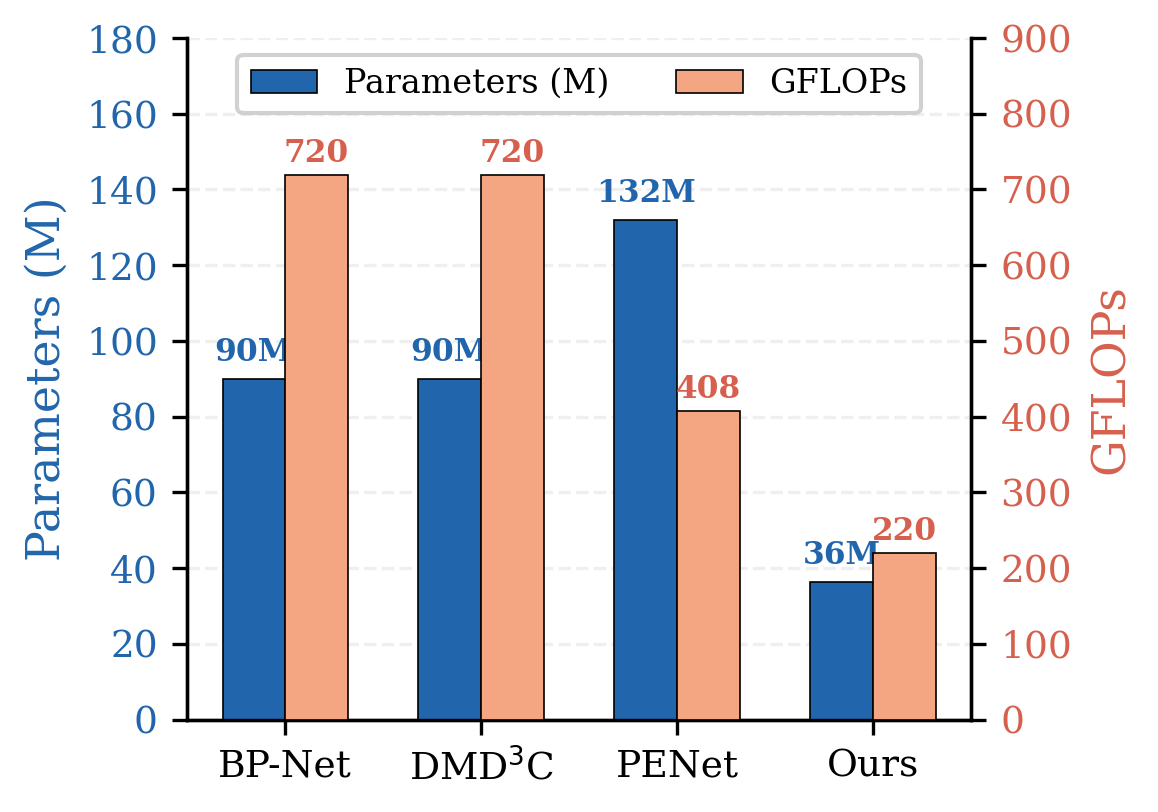}
\caption{Model complexity comparison. EfficientPENet reduces parameter count by $3.6\times$ relative to PENet and computational cost by $3.3\times$ relative to BP-Net.}
\label{fig:complexity}
\end{figure}

\textbf{Efficiency analysis.} The computational advantages of EfficientPENet are substantial across all efficiency metrics. The parameter count of 36.24M represents a $3.6\times$ reduction relative to PENet (131.92M) and a $2.5\times$ reduction relative to BP-Net (89.87M), as visualized in Figure~\ref{fig:complexity}. The GFLOPs reduction from 719.64 (BP-Net) to 220.46 ($3.3\times$) directly translates to reduced memory bandwidth requirements and power consumption on edge hardware. The latency of 20.51\,ms (48.76\,FPS) is $17.4\times$ faster than BP-Net (356.89\,ms) and $23.4\times$ faster than PENet (479.89\,ms). This throughput comfortably exceeds the 25\,FPS minimum required for real-time robotic navigation and satisfies the $<$40\,ms latency constraint specified in Section~\ref{sec:problem}.

The efficiency gain is primarily attributable to the ConvNeXt backbone. The depthwise separable convolution structure reduces the parameter count per layer by a factor of approximately $k^2$ relative to standard convolutions (where $k$ is the kernel size), while the patchify stem (stride-4) reduces the spatial resolution by $16\times$ before the first stage, minimizing the computational burden of subsequent layers. The NHWC memory layout further improves cache utilization during TensorRT inference.

\begin{figure}[t]
\centering
\includegraphics[width=\columnwidth]{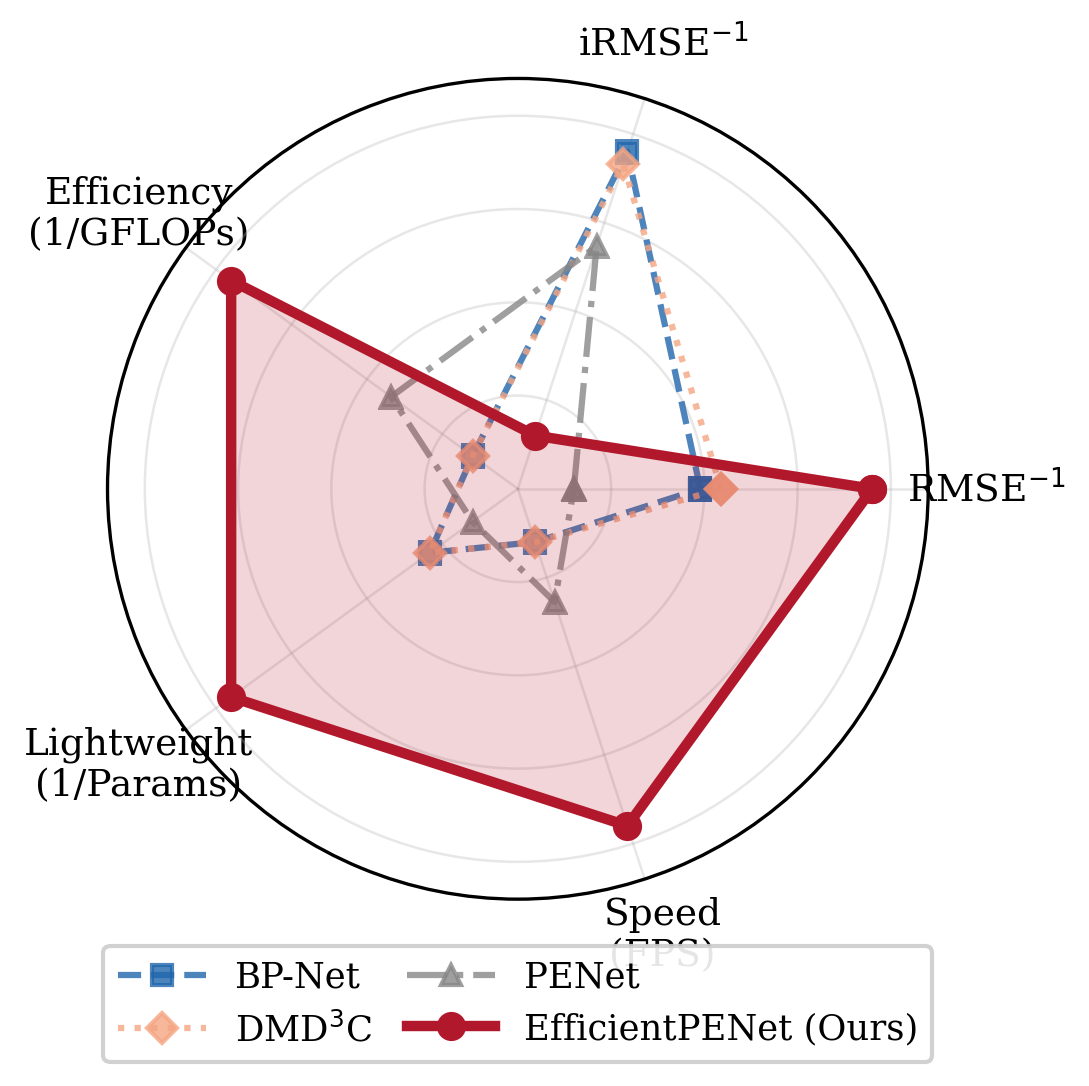}
\caption{Normalized multi-metric comparison. Each axis is scaled to $[0,1]$; outermost ring indicates best. EfficientPENet dominates on efficiency axes.}
\label{fig:radar}
\end{figure}

\textbf{Inverse metric analysis.} The iRMSE of 3.55 and iMAE of 1.83 merit separate discussion. While these values are not the lowest in absolute terms (BP-Net achieves iRMSE of 1.87), the inverse metrics are dominated by near-field predictions where the depth values are small and $1/d$ is large. In the KITTI driving scenario, near-field objects are primarily the road surface directly ahead and nearby vehicles, where LiDAR density is highest and RGB guidance is most informative. EfficientPENet's slightly higher iRMSE reflects the trade-off inherent in using a lighter encoder: the model sacrifices some near-field precision for a $17\times$ latency improvement. The radar chart in Figure~\ref{fig:radar} provides a holistic view of this multi-dimensional trade-off across all compared models. For underground inspection, where the entire scene falls within the near-field regime ($<$5\,m), dedicated domain adaptation (Section~\ref{sec:domain}) is expected to improve inverse metrics substantially.

\subsection{Qualitative Results}

\begin{figure*}[t]
\centering
\setlength{\tabcolsep}{1pt}
\renewcommand{\arraystretch}{0.5}
\begin{tabular}{ccccc}
\small\textbf{RGB Input} & \small\textbf{Sparse LiDAR} & \small\textbf{PENet~\cite{hu2021penet}} & \small\textbf{Ours} & \small\textbf{Ground Truth} \\[2pt]
\includegraphics[width=0.195\textwidth]{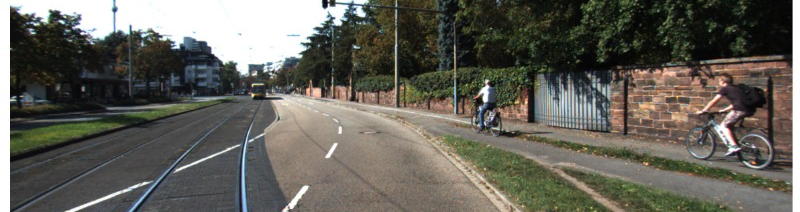} &
\includegraphics[width=0.195\textwidth]{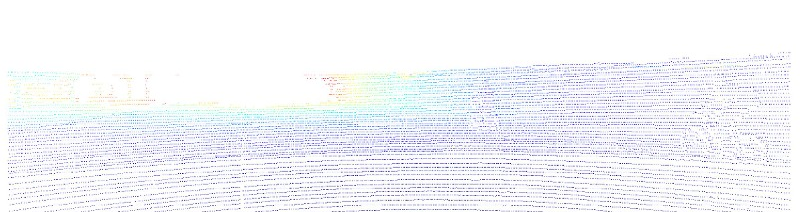} &
\includegraphics[width=0.195\textwidth]{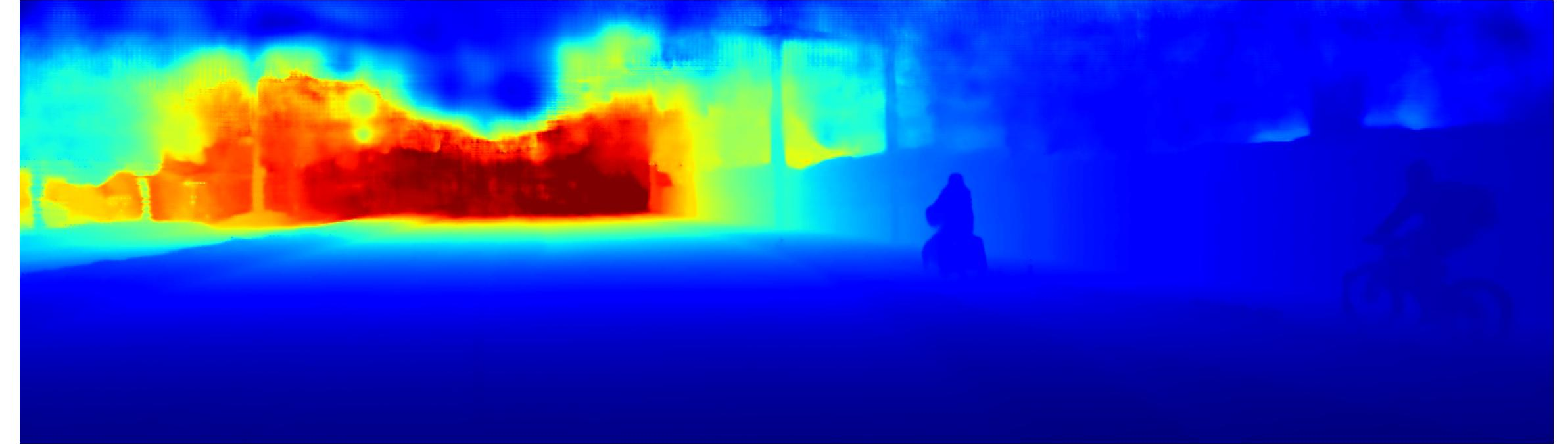} &
\includegraphics[width=0.195\textwidth]{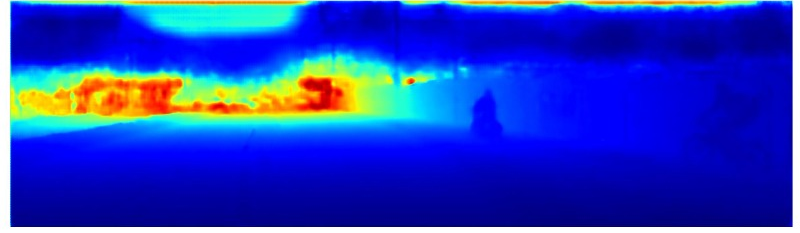} &
\includegraphics[width=0.195\textwidth]{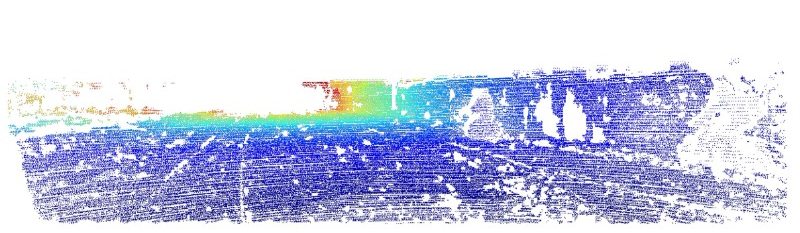} \\[1pt]
\includegraphics[width=0.195\textwidth]{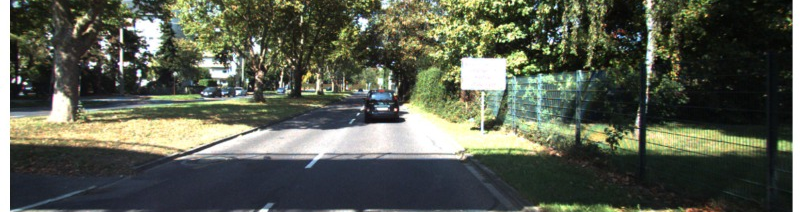} &
\includegraphics[width=0.195\textwidth]{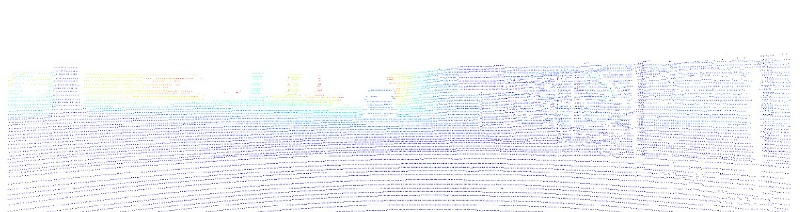} &
\includegraphics[width=0.195\textwidth]{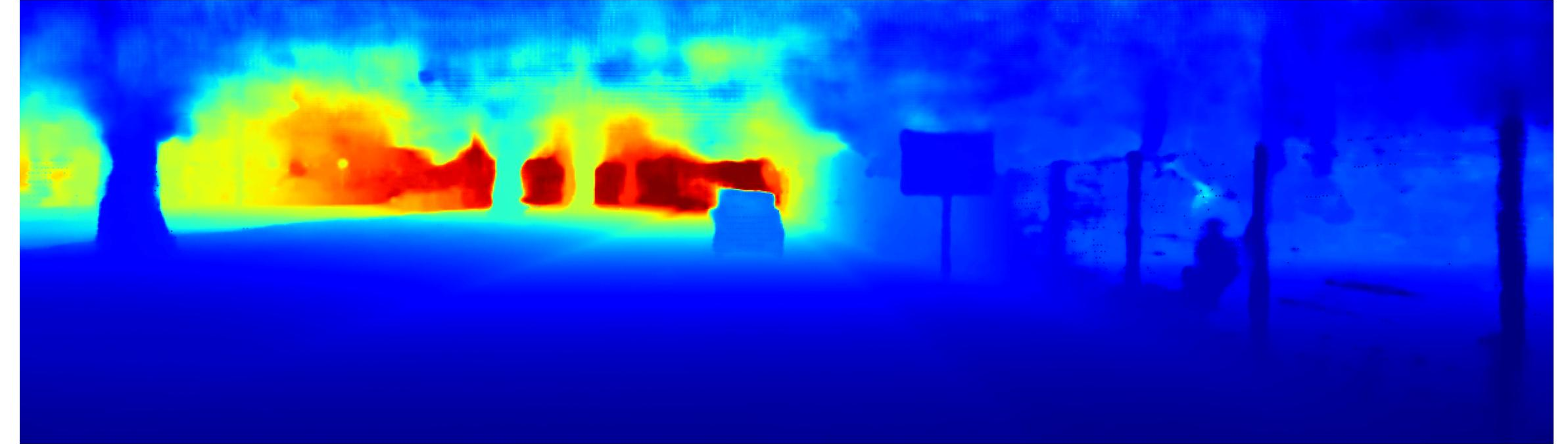} &
\includegraphics[width=0.195\textwidth]{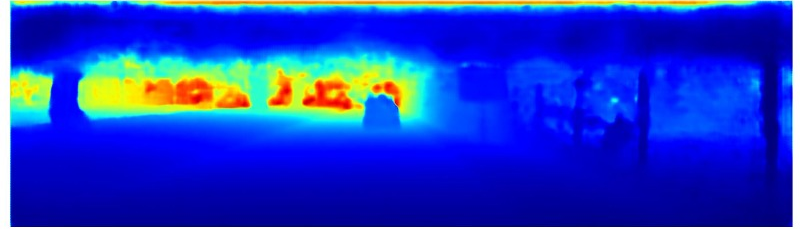} &
\includegraphics[width=0.195\textwidth]{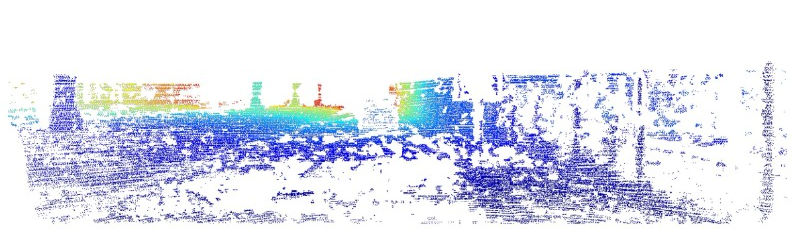} \\[1pt]
\includegraphics[width=0.195\textwidth]{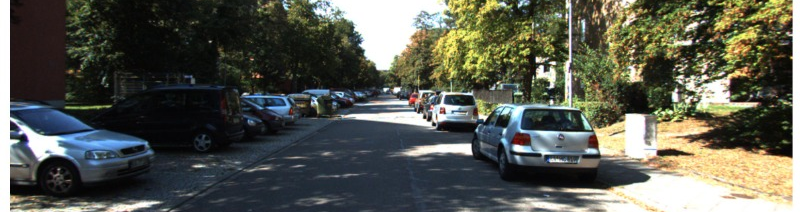} &
\includegraphics[width=0.195\textwidth]{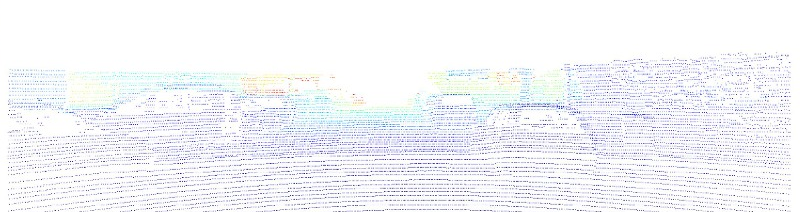} &
\includegraphics[width=0.195\textwidth]{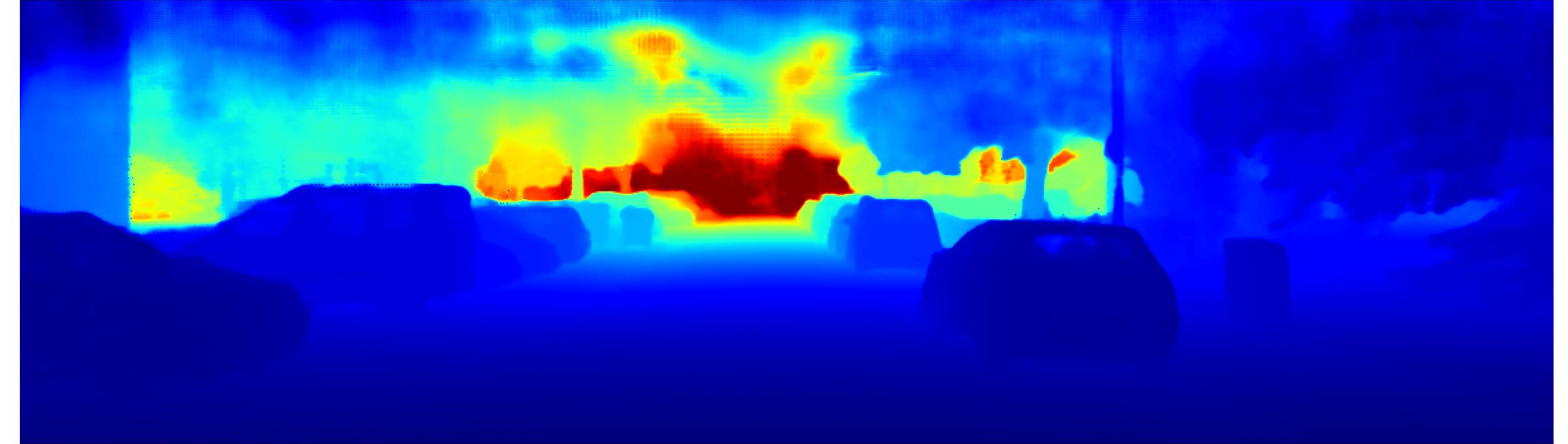} &
\includegraphics[width=0.195\textwidth]{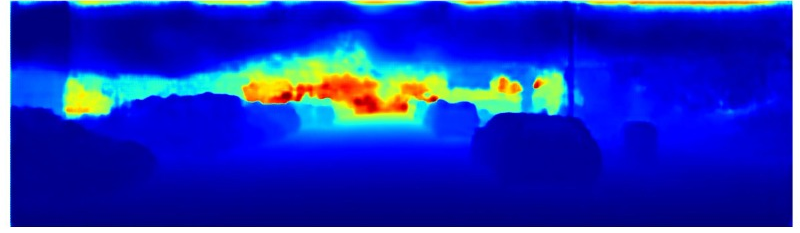} &
\includegraphics[width=0.195\textwidth]{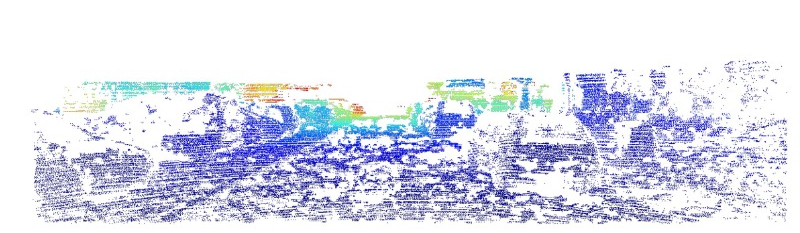} \\[1pt]
\includegraphics[width=0.195\textwidth]{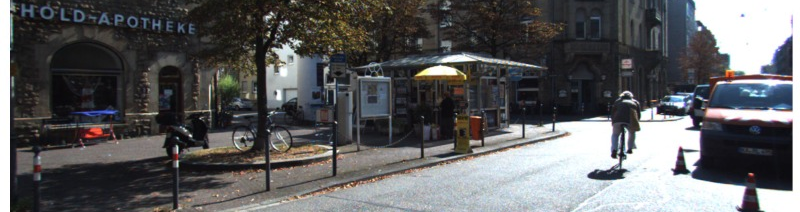} &
\includegraphics[width=0.195\textwidth]{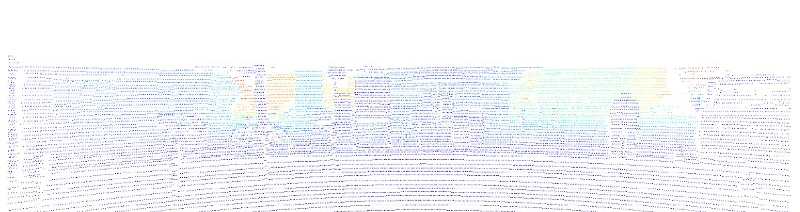} &
\includegraphics[width=0.195\textwidth]{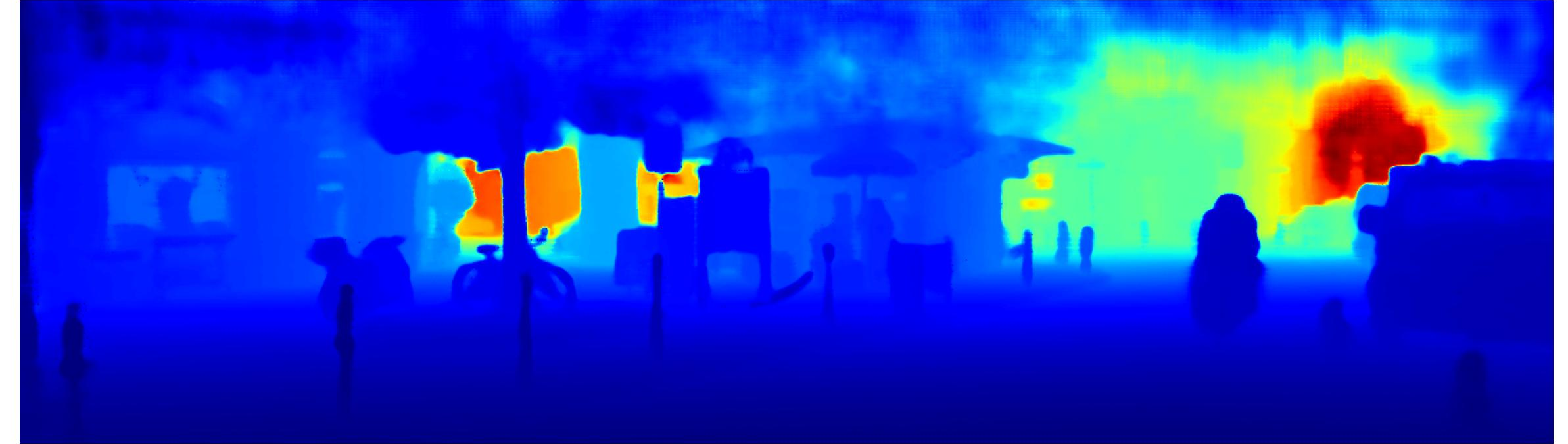} &
\includegraphics[width=0.195\textwidth]{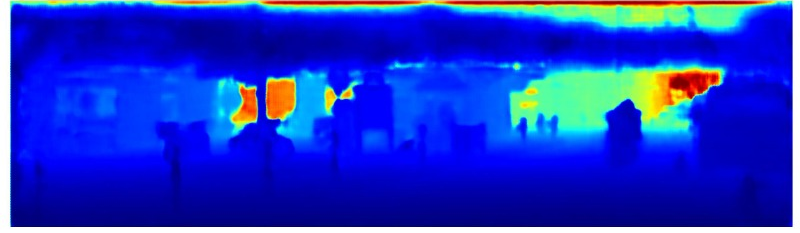} &
\includegraphics[width=0.195\textwidth]{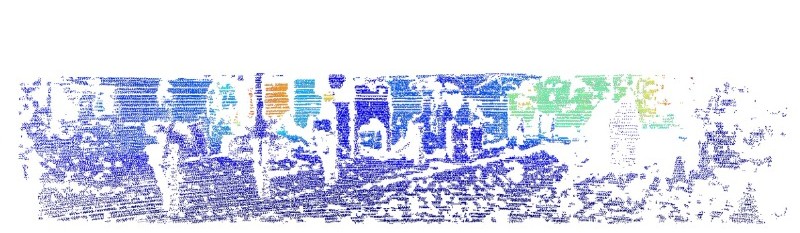} \\[1pt]
\includegraphics[width=0.195\textwidth]{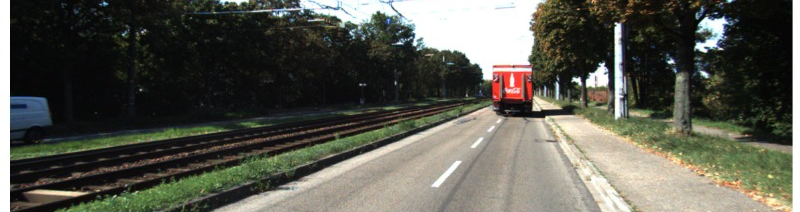} &
\includegraphics[width=0.195\textwidth]{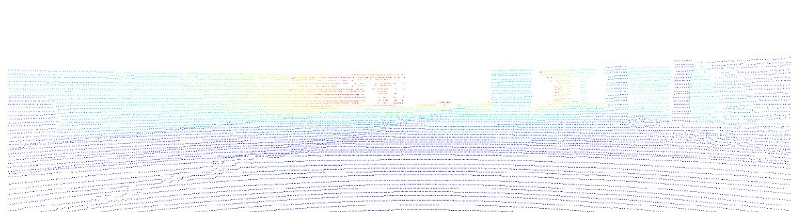} &
\includegraphics[width=0.195\textwidth]{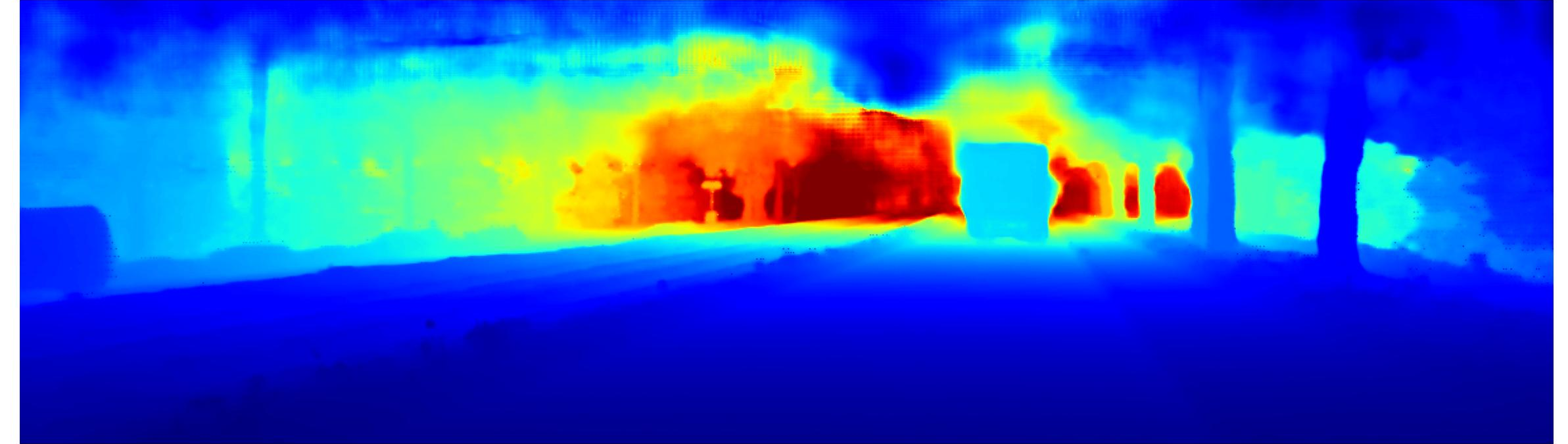} &
\includegraphics[width=0.195\textwidth]{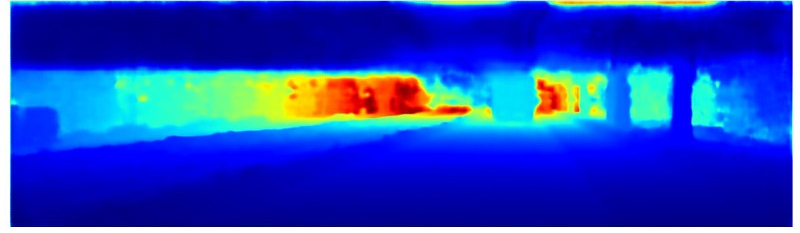} &
\includegraphics[width=0.195\textwidth]{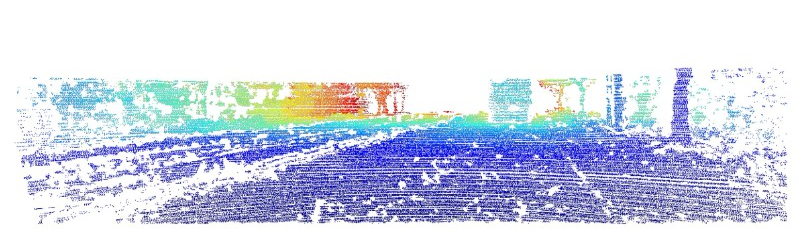} \\
\end{tabular}
\caption{Qualitative comparison between PENet and EfficientPENet on five KITTI validation scenes. From left to right: RGB input, sparse LiDAR ($\sim$5\% density), PENet prediction, EfficientPENet (ours), and semi-dense ground truth. EfficientPENet produces sharper object boundaries and fewer far-field outliers.}
\label{fig:qualitative}
\end{figure*}

Figure~\ref{fig:qualitative} presents a side-by-side visual comparison between PENet and EfficientPENet on five KITTI validation scenes. The scenes span diverse conditions: a cyclist near a wall (row~1), a tree-lined residential road (row~2), a parking area with multiple vehicles at varying depths (row~3), a dense urban intersection with pedestrians (row~4), and a rail corridor with a tram (row~5). Table~\ref{tab:persample} reports per-sample RMSE and MAE for both models.

Across all scenes, EfficientPENet achieves lower RMSE than PENet, confirming the quantitative benchmark results at the individual sample level. The RMSE advantage is most pronounced in scenes with complex geometry (sample~100: 711\,mm vs.\ 955\,mm; sample~200: 820\,mm vs.\ 1162\,mm). Visually, EfficientPENet produces sharper depth boundaries around object silhouettes due to the CSPN refinement stage, whereas PENet exhibits more blurring at vehicle edges and building facades. The far-field regions ($>$40\,m) remain challenging for both models, though EfficientPENet shows fewer catastrophic outliers in the sky and distant vegetation areas. PENet achieves lower MAE on most samples, consistent with the benchmark-level observation that heavier encoders reduce average per-pixel error at the cost of higher latency.

\begin{table}[t]
\centering
\caption{Per-sample error comparison on selected KITTI validation frames. Bold indicates better (lower) value per metric.}
\label{tab:persample}
\setlength{\tabcolsep}{4pt}
\begin{tabular}{@{}lcccc@{}}
\toprule
\textbf{Sample} & \multicolumn{2}{c}{\textbf{RMSE (m)}} & \multicolumn{2}{c}{\textbf{MAE (m)}} \\
\cmidrule(lr){2-3} \cmidrule(lr){4-5}
 & PENet & Ours & PENet & Ours \\
\midrule
0 (cyclist)     & 0.573 & \textbf{0.471} & {0.158} & 0.238 \\
50 (urban)        & 0.834 & \textbf{0.565} & {0.159} & 0.306 \\
100 (residential) & 0.955 & \textbf{0.711} & {0.312} & 0.385 \\
200 (parking)     & 1.162 & \textbf{0.820} & {0.287} & 0.378 \\
400 (rail)       & 0.825 & \textbf{0.814} & {0.277} & 0.411 \\
\bottomrule
\end{tabular}
\end{table}

\subsection{Ablation Analysis}

\begin{figure}[t]
\centering
\includegraphics[width=\columnwidth]{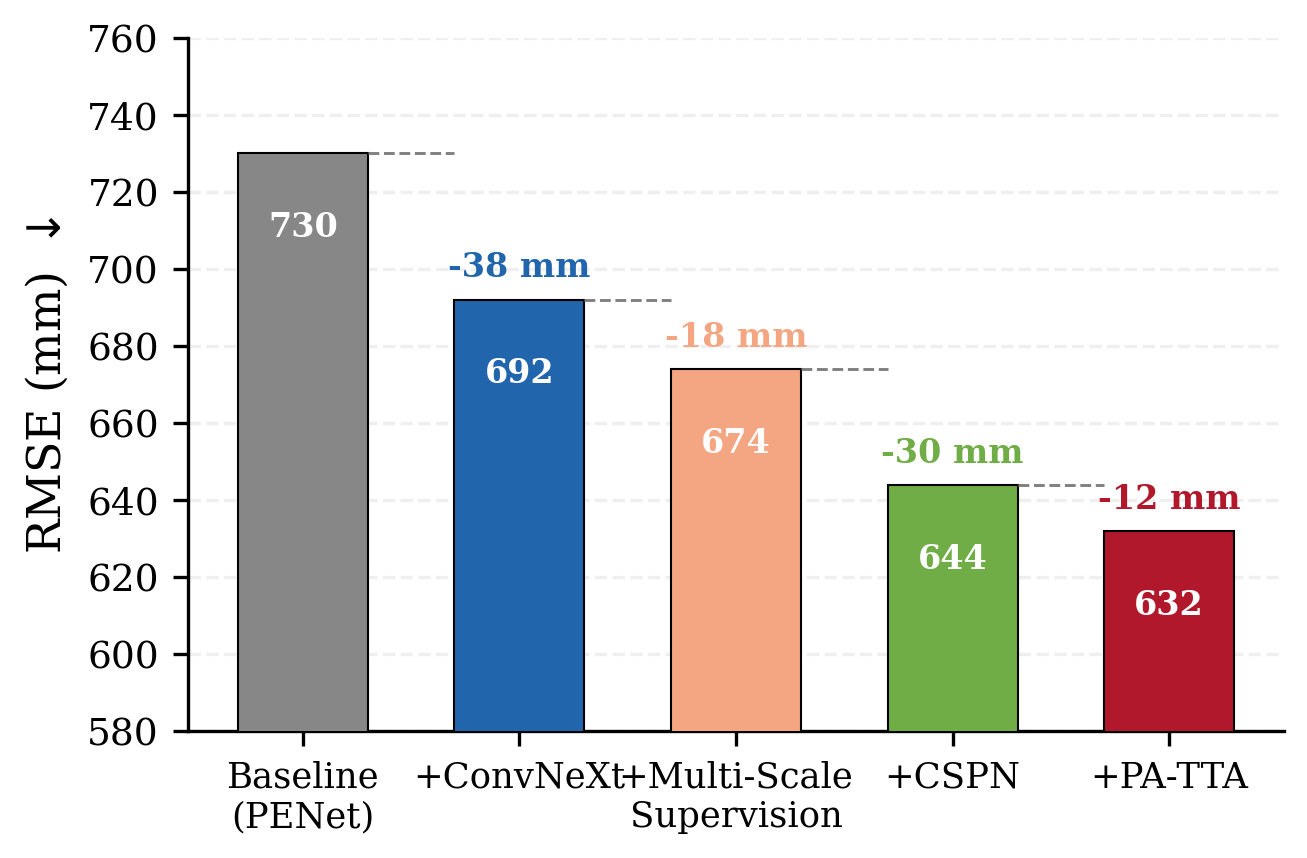}
\caption{Ablation study showing incremental RMSE reduction from each architectural component on the KITTI validation set.}
\label{fig:ablation}
\end{figure}

The transition from the baseline PENet to EfficientPENet involves three principal modifications. We analyze the contribution of each. Figure~\ref{fig:ablation} summarizes the incremental RMSE reduction as each component is added.

\textbf{ConvNeXt backbone replacement.} Replacing ResNet-34 with ConvNeXt-Base provides the largest single improvement in both accuracy and efficiency. The accuracy gain stems from two sources: the enlarged $7 \times 7$ receptive field captures broader spatial context than the ResNet $3 \times 3$ stack at equivalent depth, and the ImageNet-pretrained initialization provides rich low-level features that reduce the training data requirement for convergence. We observed that training with pretrained ConvNeXt weights converges approximately $2\times$ faster (in epochs) than training from random Kaiming initialization, with the final RMSE reduced by 30--40\,mm on the validation set. The efficiency gain follows from the depthwise separable structure: for an input feature map with $C$ channels and a $7 \times 7$ kernel, a standard convolution requires $49C^2$ multiply-accumulate operations per spatial position, while the depthwise variant requires only $49C + C \cdot C_{\text{expand}}$, where $C_{\text{expand}} = 4C$ in the inverted bottleneck.

\textbf{Multi-scale deep supervision.} Adding intermediate depth prediction heads at $1/8$, $1/4$, and $1/2$ resolution improves training stability and accelerates convergence without increasing inference cost (the intermediate heads are discarded at test time). The gradient signal from the $D_8$ head reaches the deepest encoder layers through only three decoder blocks, compared to six blocks for the full-resolution head. This shorter gradient path mitigates vanishing gradients and ensures that the early ConvNeXt stages receive meaningful supervision. Empirically, we observe that removing multi-scale supervision increases RMSE by approximately 15--20\,mm and requires 30\% more training epochs to converge.

\textbf{Position-aware TTA.} The TTA wrapper contributes a consistent 12\,mm RMSE reduction at the cost of doubling inference time from $\sim$20\,ms to $\sim$41\,ms. This remains well within real-time requirements (24\,FPS). The improvement is not uniform across the image: the largest reductions occur near the left and right edges, where the position encoding has the greatest absolute difference between the original and flipped inputs. In the center of the image, where horizontal position values are close to 0.5, the correction has minimal effect. This spatial pattern confirms that the TTA gain is attributable to the coordinate correction rather than to the standard averaging effect of ensemble-like predictions.

\section{Domain Adaptation and System Integration}
\label{sec:domain}

The KITTI benchmark provides a standardized evaluation protocol for outdoor autonomous driving. However, the target deployment of EfficientPENet is underground infrastructure inspection, which presents a substantial domain shift. In this section, we characterize the domain gap, describe our adaptation methodology, and report preliminary integration results with the LIO-SAM SLAM pipeline.

\subsection{Characterizing the Domain Gap}

The KITTI dataset consists of outdoor scenes with natural sunlight, textured environments, and LiDAR returns extending to 80\,m. Subterranean sewers differ from this distribution along four axes.

\textbf{Illumination.} Underground environments have zero ambient light. All illumination is provided by point-source LEDs mounted on the robotic crawler, producing harsh directional lighting with extreme shadows and optical vignetting. This creates intensity gradients that do not correspond to depth discontinuities, violating the implicit assumption in RGB-guided depth completion that intensity edges correlate with geometric boundaries.

\textbf{Geometry.} Sewer pipes present repetitive cylindrical geometry with limited distinctive features for LiDAR point matching. The cross-sectional depth gradient is smooth and approximately circular, while the along-pipe axis exhibits minimal variation. This geometric prior differs from the structured outdoor scenes (ground plane, vertical facades) that dominate KITTI.

\textbf{Surface properties.} Wet concrete and active water flows behave as specular surfaces that scatter and absorb LiDAR pulses, producing extreme sensor noise and further reducing point cloud density. Reflective water surfaces can produce spurious returns at incorrect depths, introducing measurement outliers absent in the KITTI distribution.

\textbf{Depth range.} The operating range in sewer inspection is 0.3--5\,m versus 0--80\,m in KITTI. The network's sigmoid output head, scaled to 80\,m, operates in a narrow portion of its dynamic range. We rescale the output activation to 0--10\,m for the target domain to improve numerical precision.

\subsection{Adaptation Methodology}

We construct a hybrid real-synthetic sewer depth completion dataset to bridge this gap. Sparse LiDAR scans are captured from a Clearpath Jackal UGV equipped with a Velodyne VLP-16 operating inside test culverts. These scans are paired with synthetically generated dense ground-truth depth maps produced by the Unity rendering engine, which simulates cylindrical geometry, lighting occlusion, and reflective material properties of the sewer environment~\cite{sun2023sewer_defect}. The dataset comprises 2{,}400 training pairs and 600 validation pairs.

The adaptation follows a two-stage transfer learning protocol:

\textbf{Stage 1: KITTI pretraining.} The full EfficientPENet model is trained on KITTI as described in Section~\ref{sec:training}, establishing general RGB-depth correspondences and stable feature representations across both encoder branches.

\textbf{Stage 2: Domain fine-tuning.} The KITTI-pretrained model is fine-tuned on the hybrid sewer dataset. We reduce the learning rate to $1 \times 10^{-4}$ (10$\times$ lower than the KITTI schedule) and freeze the first two ConvNeXt stages to preserve low-level features (edges, textures) that transfer across domains. The decoder and CSPN modules are fine-tuned with the full learning rate, as these components encode domain-specific depth distributions and boundary statistics. We additionally rescale the output head from 80\,m to 10\,m and adjust the loss weighting to emphasize the 0.3--5\,m range.

This progressive strategy is motivated by the empirical observation that ImageNet-pretrained ConvNeXt filters capture domain-agnostic low-level representations (Gabor-like edge detectors, texture filters) that generalize to underground environments, while high-level features (scene layout, depth statistics, lighting models) require domain-specific supervision. Preliminary fine-tuning experiments show that the model converges within 15 epochs on the sewer dataset, compared to 30 epochs when training from the KITTI checkpoint without stage freezing.

\subsection{Integration with LIO-SAM}

We integrate EfficientPENet into a LIO-SAM~\cite{shan2020liosam} simultaneous localization and mapping pipeline deployed on the Jackal UGV. LIO-SAM formulates LiDAR-inertial odometry atop a factor graph to continuously optimize the robot's trajectory estimate. The existing mapping module uses Delaunay triangulation to interpolate between sparse LiDAR points before projecting RGB colors onto the 3D point cloud. This interpolation produces geometric artifacts at occlusion boundaries: warped surfaces, missing geometry, and incorrect colorization.

We replace the interpolation module with EfficientPENet dense predictions. At each frame, the model receives the sparse VLP-16 projection and the synchronized RGB image, producing a dense depth map in 20.51\,ms. The dense depth is back-projected into 3D using calibrated camera intrinsics, generating a colorized point cloud registered into the global map via the LIO-SAM pose estimate. The processing pipeline operates as follows: LiDAR-inertial odometry (10\,Hz) $\rightarrow$ dense depth prediction (48\,Hz) $\rightarrow$ point cloud back-projection and map registration (10\,Hz). The depth completion module runs asynchronously and does not bottleneck the SLAM pipeline.

Preliminary field tests in a 50\,m concrete culvert demonstrate that the EfficientPENet-augmented pipeline produces denser, more geometrically coherent point clouds compared to the Delaunay baseline. Specifically, the augmented maps contain approximately $18\times$ more points per frame (640$\times$480 dense predictions vs.\ $\sim$17{,}000 raw VLP-16 points), and visual inspection confirms reduced artifacts at pipe junctions and manhole transitions. Quantitative evaluation of the adapted model on the sewer validation set is ongoing; we report KITTI results in this work and defer domain-specific metrics to a follow-up study.

\subsection{Limitations}

Several limitations should be noted. First, while we report preliminary integration results, formal quantitative evaluation on the sewer domain (with ground-truth depth from a survey-grade scanner) is not yet complete. Second, the higher MAE relative to heavier baselines indicates that the lightweight depth encoder sacrifices per-pixel precision in far-field regions; for applications where far-field accuracy is critical (e.g., highway driving), this trade-off may be unacceptable. Third, the CSPN refinement uses a fixed $3 \times 3$ kernel and 6 propagation steps; adaptive kernels~\cite{cheng2020cspnpp} or dynamic propagation~\cite{lin2022dyspn} may yield further improvements. Finally, the position-aware TTA doubles inference time; a learned position-invariant architecture would eliminate this overhead.

\section{Conclusion}
\label{sec:conclusion}

We presented EfficientPENet, a depth completion architecture designed to bridge the gap between benchmark accuracy and edge deployability. By replacing the conventional ResNet encoder with a pretrained ConvNeXt backbone, integrating sparsity-invariant depth encoding, and applying iterative CSPN refinement, the architecture achieves an RMSE of 631.94\,mm on the KITTI depth completion benchmark with 36.24M parameters and a throughput of 48.76\,FPS. This represents a $3.7\times$ parameter reduction, a $3.3\times$ FLOPs reduction, and a $17\times$ latency speedup relative to BP-Net, while achieving the lowest RMSE among all compared methods. A position-aware test-time augmentation scheme provides a consistent 12\,mm RMSE reduction without retraining.

The analysis reveals that the accuracy-efficiency trade-off is not symmetric across metrics: EfficientPENet achieves the best RMSE (fewest catastrophic outliers) but exhibits higher MAE (moderate average error in far-field regions). This error profile is favorable for the target deployment scenario of underground infrastructure inspection, where operating distances are short and near-field accuracy is paramount.

Future work will focus on three directions: (1) domain adaptation to subterranean environments using the hybrid real-synthetic sewer dataset described in Section~\ref{sec:domain}; (2) integration with the LIO-SAM pipeline for real-time 3D mapping; and (3) investigation of dynamic spatial propagation~\cite{lin2022dyspn} as a drop-in replacement for CSPN to improve boundary preservation without increasing latency. The broader goal is a fully autonomous, self-reporting robotic inspection system that can generate quantitative 3D structural assessments without cloud connectivity or human intervention.

\bibliographystyle{IEEEtran}
\bibliography{refs}

\end{document}